\documentclass[runningheads]{llncs}
\usepackage{subfig}
\usepackage{hyperref}
\usepackage{graphicx}
\usepackage{amssymb}
\usepackage{amsmath}
\usepackage[ruled,vlined]{algorithm2e}
\SetKwInput{KwInit}{Initialize}
\SetKwInput{KwProc}{Procedure}

\begin{document}
\title{Person Detection Using an Ultra Low-resolution Thermal Imager on a Low-cost MCU}

\titlerunning{Person Detection Using an Ultra Low-resolution Thermal Imager}
%
\author{First author\inst{1}\orcidID{0000-1111-2222-3333} \and
Second author\inst{1}\orcidID{0000-1111-2222-3333} \and
Third author\inst{1}\orcidID{0000-1111-2222-3333} \and
Fourth author\inst{1}\orcidID{0000-1111-2222-3333}}
\authorrunning{F. author et al.}

\author{Maarten Vandersteegen\inst{1}\orcidID{0000-0002-9377-0922} \and
Wouter Reusen\inst{2} \and
Kristof Van Beeck\inst{1}\orcidID{0000-0002-3667-7406} \and
Toon Goedem\'e\inst{1}\orcidID{0000-0002-7477-8961}}
\authorrunning{M. Vandersteegen et al.}

%

\institute{KU Leuven EAVISE, Jan Pieter De Nayerlaan 5, Sint-Katelijne-Waver, Belgium \\
\email{\{firstname.lastname\}@kuleuven.be} \and
Melexis Technologies NV, Transportstraat 1, Tessenderlo Belgium \\
\email{wre@melexis.com}}
\maketitle              
\begin{abstract}
Detecting persons in images or video with neural networks is a well-studied subject in literature. However, such works usually assume the availability of a camera of decent resolution and a high-performance processor or GPU to run the detection algorithm, which significantly increases the cost of a complete detection system. However, many applications require low-cost solutions, composed of cheap sensors and simple microcontrollers. In this paper, we demonstrate that even on such hardware we are not condemned to simple classic image processing techniques.  We propose a novel ultra-lightweight CNN-based person detector that processes thermal video from a low-cost 32$\times$24 pixel static imager. Trained and compressed on our own recorded dataset, our model achieves up to 91.62\% accuracy (F1-score), has less than 10k parameters, and runs as fast as 87ms and 46ms on low-cost microcontrollers STM32F407 and STM32F746, respectively.

\keywords{Person Detection \and Low-resolution \and Thermal \and Neural Networks \and Microcontrollers \and Pruning \and Quantization}
\end{abstract}
\section{Introduction}

\begin{figure}
    \centering
    \includegraphics[width=\textwidth]{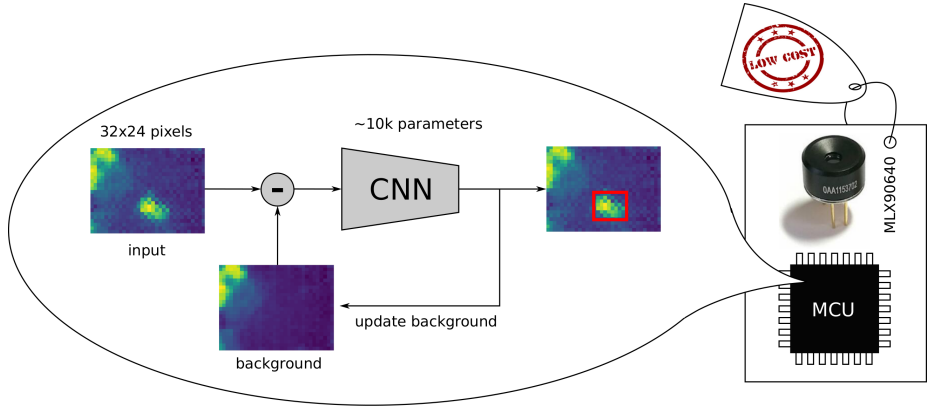}
    \caption{Our proposed system is built upon a low-cost thermal imager (MLX90640) and a low-cost MCU, which runs a custom designed and compressed CNN object detector with a unique background subtraction mechanism.}
    \label{fig:overview}
\end{figure}

Bounding-box based person detection using neural networks is one of the most covered topics in computer vision literature \cite{ren2015faster,redmon2017yolo9000,liu2016ssd,law2018cornernet} which is a strong indication of its importance for many applications. However, most studies assume moderate to high quality sensors and powerful processors like GPUs with lots of resources to work with, which comes with bulky hardware, high power consumption and a big price tag in the order of thousands of dollars. Moreover, high quality sensors such as color cameras may reveal privacy-sensitive information about persons, leading to privacy concerns and mistrust in technology.
In many real-life applications, such problems are unacceptable and therefore low-resolution sensors and embedded GPUs, low-power accelerators or even standard microcontrollers (MCUs) are preferred, costing rather hundreds of dollars, tens of dollars or even a few dollars, respectively. With an exponential decrease in price, comes an exponential decrease in computational resources, which makes deployment of neural networks increasingly complex, especially for MCUs.
This requires smart solutions that work with low-resolution sensors, sparse model architectures and state-of-the-art compression technologies to squeeze out every last redundant computation, which in turn poses challenges in retaining sufficient prediction accuracy.

Nonetheless, this work succeeds in building a highly accurate person detection system with a low-cost MCU, the cheapest category of processors, a low-cost sensor and a highly optimized deep-learning based detection algorithm.
Due to the thermal spectrum, combined with ultra low-resolution, it must be noted that the identity of persons in front of the camera is guaranteed to be preserved \cite{kraft2021low}.

\begin{figure}
    \centering
    \includegraphics[width=\textwidth]{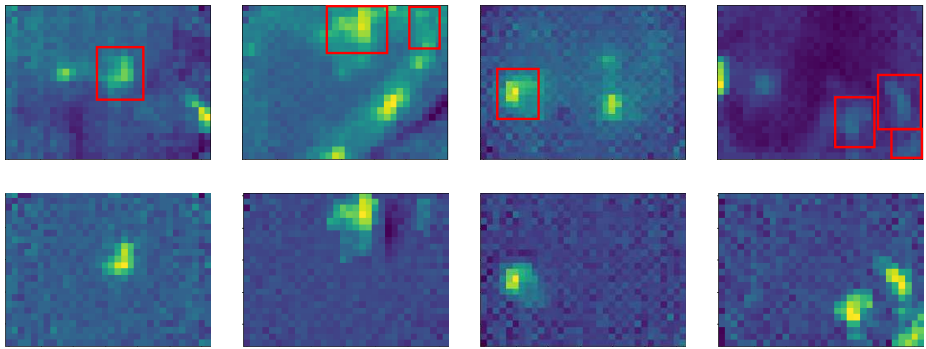}
    \caption{Example images from the dataset with annotated persons (top row) and the same images with our proposed background subtraction applied (bottom row), which masks out static objects that seem indistinguishable from a real person.}
    \label{fig:with_and_without_bg_sub}
\end{figure}

Our methodology, depicted in Figure \ref{fig:overview}, consists of a custom CNN-based single-stage object detector and a unique video background subtraction algorithm to help distinguish static person-like objects from real persons.
Given the fact that deep learning is skilled at extracting useful context from the background of an image, the combination of deep learning and background subtraction might seem odd at first glance. However, this is not the case when working with such low-resolution (thermal) images, since background features from our sensor are often indistinguishable from a person, even for a human observer. Figure \ref{fig:with_and_without_bg_sub} depicts a few of such difficult examples, with and without our background subtraction applied.

Our contributions are summarized as follows:
\begin{itemize}
    \item {This work proposes a truly low-cost and privacy preserving detection system for stationary setups, featuring a novel detection algorithm.}
    \item {We introduce a benchmarking system to compare the accuracy of our model against the baseline -- the standard person detection software from Melexis, the manufacturer of the MLX90640 thermal sensor -- and prove that our models outperform the former, achieving up to 15.6\% higher in $F_1$-score.}
    \item {We are the first to prove the advantage of background-subtraction in combination with a CNN for ultra low-resolution thermal object detection.}
    \item {We present extensive compression experiments and a benchmark on inference time and memory utilization of our models deployed with Tensorflow Lite or microTVM on two MCU targets.}
    \item {We release our dataset in order to support future research\footnote{Our dataset: \url{https://iiw.kuleuven.be/onderzoek/eavise/mldetection/home}}.}
\end{itemize}

Our paper is structured as follows: Section \ref{sec:related work} discusses related work on person detection and neural network compression, given their direct relevance to our application, Section \ref{sec:approach} elaborates on our detection algorithm, the proposed compression method and how we collected our dataset, Section \ref{sec:experiments} discusses our results while conclusions are made in final Section \ref{sec:conclusion}.

\section{Related Work}
\label{sec:related work}

\subsection{Person Detection}

Since the dawn of the deep-learning area, person detection has been addressed mostly with neural network based object detectors such as (Fast(er)) R-CNN \cite{ren2015faster}, YOLO \cite{redmon2017yolo9000} and SSD \cite{liu2016ssd}. In a constant battle to improve the accuracy and reducing the computational budget, many new designs and improvements emerged in later years \cite{lin2017focal,law2018cornernet,tan2020efficientdet,wong2019yolo}. Other attempts for improving the accuracy focus on (1) exploiting temporal queues in video object detection and (2) using different sensor modalities.

Video object detection or recognition methods exploit the temporal domain to improve the detection accuracy. Liu et al. \cite{liu2018mobile} for example, propose to use a convolutional LSTM module into their SSD object detector, Simonyan et al. \cite{simonyan2014two} use an additional optical-flow input image in their two-stream approach, Li et al. \cite{li20183d} experiment with 3D convolutions in their video-based vehicle detector, and Kang et al. \cite{kang2017t} use a tubelet tracking mechanism to improve their detection results.

Different sensor modalities are incorporated in several works for more robust detection in difficult viewing circumstances. Experiments with time-of-flight cameras \cite{xiang2021real}, thermal cameras \cite{ippalapally2020object,jiang2022object} or a combination of different sensors \cite{vandersteegen2018real,wolpert2020anchor,ophoff2019exploring} can be found in the literature.

Although low-resolution object detection with deep-learning has been attempted down to 96x96 pixels \cite{callemein2020low}, or even lower with the help of super-resolution \cite{haris2021task}, we are the first to try this on 32$\times$24 pixels in a direct way. A method for counting people in a detection-like way with the same sensor as ours \cite{kraft2021low} exists, however their dataset seems less challenging (no person-like background objects) and their model is much heavier (130k params) compared to ours (10k params).

Constrained by the very limited resources of regular MCUs, we base our model on tiny YOLOv2 \cite{redmon2017yolo9000}, which is small and much more scalable compared to two-stage detectors or multi-headed detectors like SSD or later YOLO versions.
Even though we process video, spatio-temporal building blocks with a large memory footprint like 3D convolutions are to be avoided, together with convolutional LSTMs, given that the latter are difficult to tune and make compression extremely complex. We however included experiments with a motion image \cite{simonyan2014two} in Section \ref{sec:results_bg_sub}, because it's cheap to calculate.

\subsection{Model Compression}

Although several efficient CNN architectures like MobileNet \cite{howard2017mobilenets} or EfficientNet \cite{tan2019efficientnet} emerged from the need for lighter models, additional compression is often required to meet model size, computational budget, energy, or time constraints.
The most popular compression techniques can be divided in three categories: (1) model quantization, (2) model pruning and (3) Network-Architecture-Search (NAS).

Quantization approaches come in two flavors: (1) Post-Training Quantization (PTQ) and (2) Training-Aware Quantization (QAT). PTQ is the easiest and most widely used quantization approach for obtaining 8-bit CNNs and is a standard feature in most deployment frameworks like Tensorflow Lite \cite{tflite} or TensorRT \cite{tensorrt}, while QAT is the preferred choice for sub-8-bit precision \cite{nagel2021white}.

Pruning methods can be divided in two categories: (1) unstructured pruning and (2) structured pruning. Unstructured pruning methods \cite{lecun1989optimal,han2015learning,laurent2020revisiting} aim to remove unnecessary neural connections by setting individual weight values to zero. However, this introduces weight sparsity which requires special libraries to support the acceleration of such models. Structured pruning methods like filter pruning \cite{li2016pruning,liu2017learning,he2018soft,ophoff2021investigating} on the other hand, remove entire convolution filters at once and avoid the need for special acceleration libraries or hardware.
Different methods are proposed to identify good filter candidates for pruning, ranging from methods that remove filters with the smallest $L_1$ or $L_2$ filter norms \cite{li2016pruning,he2018soft,ophoff2021investigating}, to more complex ones that for example require changes in the loss function \cite{liu2017learning}.

NAS has been adopted by several works \cite{banbury2021micronets,lin2020mcunet,hendrickx2022hot} to find small architectures that can be directly deployed on resource constrained devices. Although such techniques show remarkable performance, their complexity is not needed in our work. Instead, we prefer iterative channel pruning with simple norm-based saliency, which has been proved by Ophoff et al. \cite{ophoff2021investigating} to work surprisingly well for constrained object detection problems.
We believe that our approach can be defined as a constrained object detection problem because: (1) we only have a single class, (2) we target fixed camera viewpoints and (3) most importantly, our background subtraction method greatly reduces the background variance in the input. These constraints allow simple, but well proven structured pruning methods, to compress our model by more than a factor $\times$100. In addition, we also quantize our models to 8-bit through regular PTQ.

\section{Approach}
\label{sec:approach}

In order to train a CNN-based person detector, a large scale dataset is needed. Since no person detection datasets of our target sensor (MLX90640) are publicly available, we recorded and annotated our own dataset, which is described in more detail in Section \ref{sec:approach_dataset}. Section \ref{sec:approach_model} describes our CNN-based detector and its background subtraction technique, and Section \ref{sec:approach_compression} elaborates on the proposed model compression method.

\subsection{Dataset}
\label{sec:approach_dataset}

\vspace{-3em}
\begin{figure}
    \centering
    \subfloat[]{\includegraphics[width=.45\textwidth]{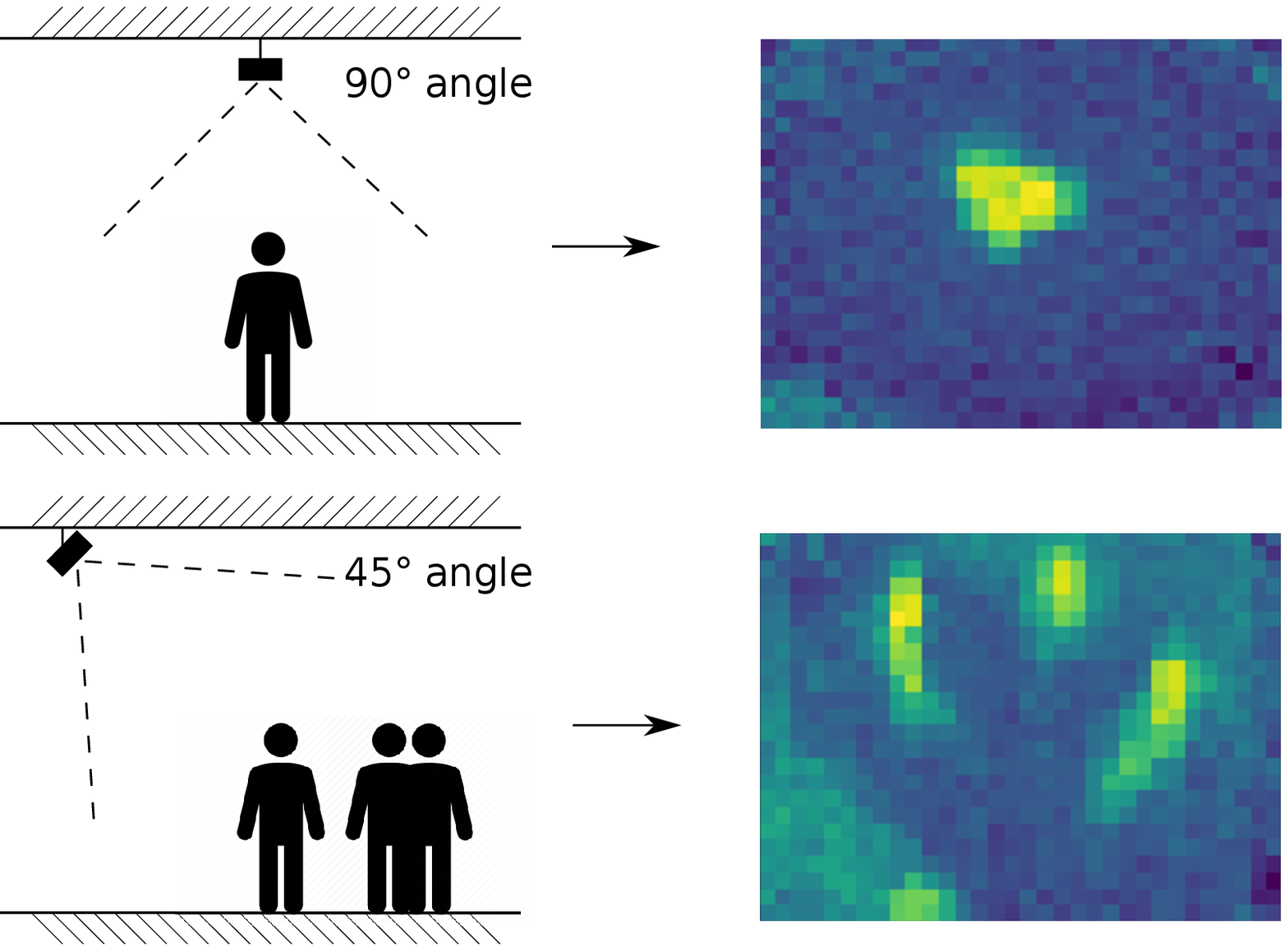}
    \label{fig:recording_angles}}
    \quad
    \subfloat[]{\includegraphics[width=.45\textwidth]{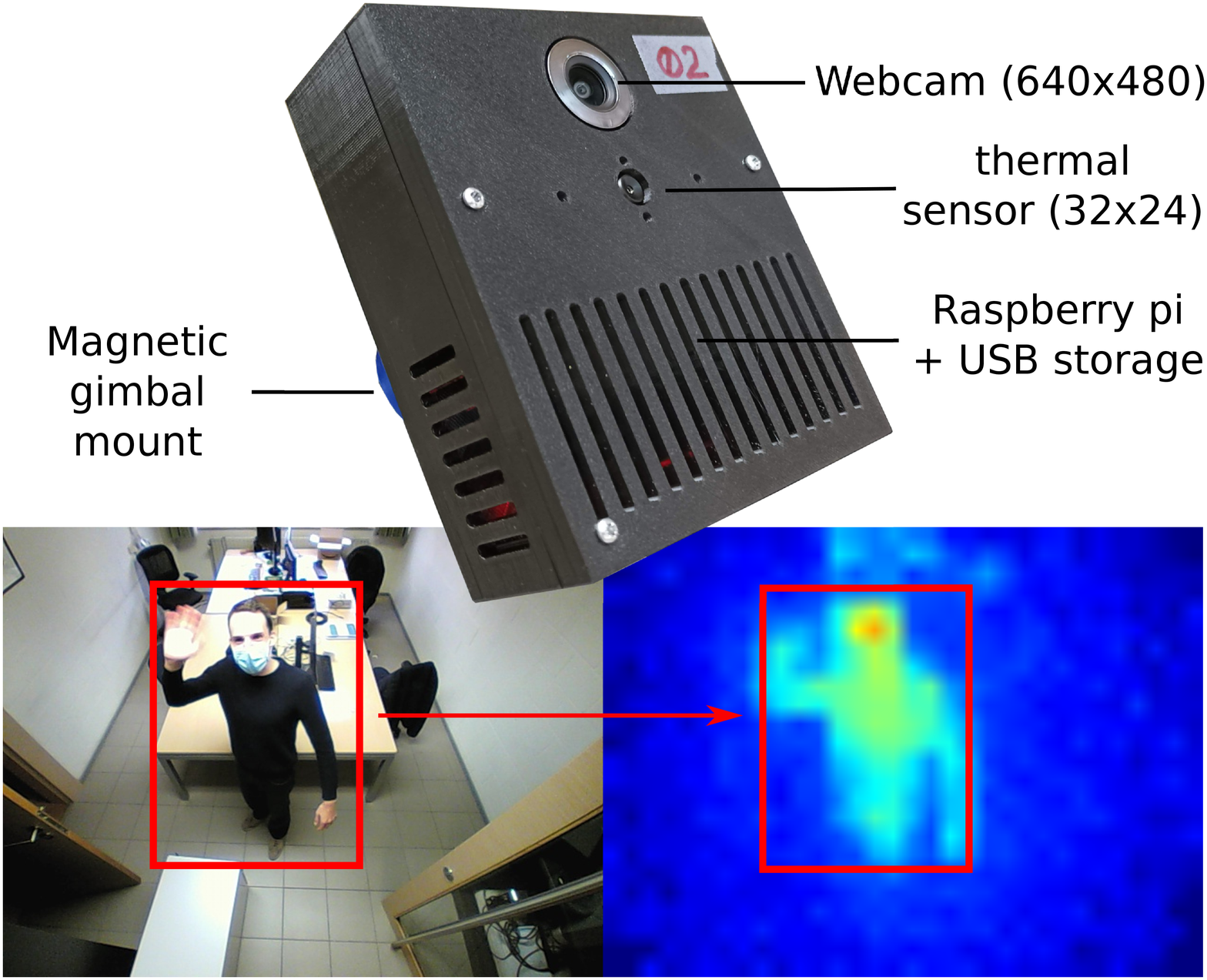}
    \label{fig:dataset_recorder}}
    \caption{(a) Illustration of the 45-degree and 90-degree recording angles maintained in our dataset and (b) our recorder featuring the MLX90640 thermal sensor and a webcam. After running existing tracking software on the visible video, the bounding boxes can be used as annotations for the thermal images after manual inspection.}
\end{figure}

A large bounding-box annotated dataset is constructed of 190 lengthy video clips at 8FPS, containing 96k thermal video frames in total. Recordings are made using a static camera setup that is mounted on the ceiling in 12 different locations including offices, residential rooms and laboratory rooms. Each video clip is labeled with a \textit{45-degree} or \textit{90-degree} tag, indicating that the setup is pointed straight down or at an angle of approximately 45 degrees, respectively, as shown in Figure \ref{fig:recording_angles}.
Video frames are stored as 16-bit .tiff images containing the raw temperature measurements in degrees Celsius, multiplied by 100.
We split our dataset in train, validation and test set, where the videos of the validation and test sets are recorded at different locations compared to the videos of the training set, in order to avoid overfitting.



To help automate the annotation process, existing person detection software, based on Mask-RCNN \cite{he2017mask} and an object tracker, is used to track persons in a high resolution color video feed, coming from a webcam that is included in the recorder setup. Since the recording angle and field-of-view of both cameras have been made equal, the generated tracking data from the webcam can be used as annotation data for the thermal video, after careful manual inspection and correction of the tracking data using the CVAT \cite{cvattool} annotation tool. Figure \ref{fig:dataset_recorder} depicts our recorder and an observation example of both sensors.

\subsection{Model}
\label{sec:approach_model}

\vspace{-2em}
\begin{figure}
    \centering
    \includegraphics[width=\textwidth]{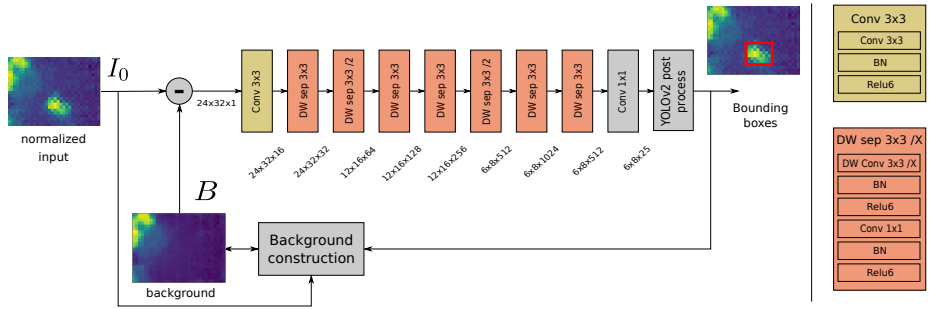}
    \caption{Our proposed model is a CNN based person detector which infers from a background subtracted input image and a differential motion image.}
    \label{fig:model}
\end{figure}

Our proposed model, illustrated in Figure \ref{fig:model}, consists of a simple CNN with a YOLOv2 detection head and a background subtraction algorithm.  First, a constructed background image $B \in \mathbb{R}^{H\times W}$ is subtracted from the current normalized input image $I_0 \in \mathbb{R}^{H\times W}$. Second, the result is sent through the CNN backbone, consisting of a single 3$\times$3 convolution followed by 7 depthwise-separable convolutions. Finally, the YOLOv2 head, a 1$\times$1 convolution followed by its typical post-processing steps, produces the bounding boxes. The following subsections (1) motivate the addition of a background subtraction algorithm, while discussing its implementation and (2) elaborate on the design choices of our CNN architecture.

\subsubsection{Background Subtraction.}
This allows the model to make a clear distinction between stationary objects with a person-like heat signature and actual persons. Distinguishing a person from a computer for example on a single 32$\times$24 pixel thermal image, is often impossible without further context, even for a human observer. The top row images from Figure \ref{fig:with_and_without_bg_sub} depict a few of these difficult samples from our dataset to illustrate this. In such cases, the only difference between both objects is that a person moves from time to time, while a stationary object does not. Based on this assumption, a robust distinction can be made by maintaining an up-to-date background image of stationary features. Our experiments in Section \ref{sec:results_bg_sub} prove that this method gives an additional boost of up to 28\% in AP detection accuracy.

Our proposed background updating algorithm, given by Algorithm \ref{alg:background_update}, uses the current input image $I_0$, the current background image $B$ and the detections $D$ from the current input image to produce the updated background image $B^*$. First, all bounding boxes in $D$ are enlarged by one pixel in all directions.
Second, a binary mask $M$ is constructed that contains white pixels at locations that overlap with one or more bounding boxes from $D$, and black pixels otherwise. Third, a background candidate $\hat{B}$ is created from background pixels of $I_0$ and foreground pixels of $B$, and finally the new background $B^*$ is created through an exponential-moving-average filter with decay factor $\alpha = 0.99$. Figure \ref{fig:with_and_without_bg_sub} shows the qualitative effect of our background subtraction algorithm on a few samples from our dataset.

\begin{algorithm}
    \SetAlgoLined
    \KwIn{input image $I_0 \in \mathbb{R}^{H\times W}$, current background image $B \in \mathbb{R}^{H\times W}$ and detections bounding boxes $D$}
    \KwOut{updated background image $B^*$}
    \KwProc{ \\
        $D \leftarrow \text{enlarge\_boxes}(D)$ \\
        $M \in \{0, 1\}^{H\times W}$ \\
        $\forall i \in \{1,...,H\}, \forall j \in \{1,...,W\}$ \\
        $\quad M_{i,j} =
        \begin{cases}
            1 \quad \text{if } i,j \text{ overlaps with a bounding box from } D\\
            0 \quad \text{otherwise}
        \end{cases}$ \\
        $\hat{B} \leftarrow B \odot M + I_0 \odot (1 - M)$ \\
        $B^* \leftarrow  \alpha B + (1 - \alpha) \hat{B}$
    }
    \caption{Background update}
    \label{alg:background_update}
\end{algorithm}

During initialization, the first background image is constructed by averaging three consecutive frames, in the assumption that no persons are present at startup. Additional computations for calculating the background are negligible, since this happens only once every 25 frames.
During training, ground truth annotations are used to construct the background image instead of detections.

\subsubsection{Network Architecture.}
Our network architecture, is inspired by tiny YOLOv2 because we value some of its design choices: (1) a single headed output, which simplifies the network graph and its post-processing, (2) a small amount of layers, which significantly reduces the time overhead for calling the layers, and (3) no residual connections, other branching structures or exotic layer types to simplify compression and deployment. Moreover, the number of layers and number of output channels per layer of our uncompressed model are the same as in tiny YOLOv2 (details in Figure \ref{fig:model}). We however prefer depthwise-separable convolutions in order to create a sparser model to start with, therefore reducing the burden on the compression stage. ReLU6 is preferred over leaky ReLU since it can be fused into its preceding convolution layer in most deployment frameworks.
To retain sufficient spatial output resolution, our model down-samples the input resolution by a factor four, in contrast to the more commonly used factor 32 in other higher resolution object detectors. Our model has $num\_anchors \times 5 = 5 \times 5 = 25$ output channels with anchor box sizes that are adjusted to our dataset.

\subsection{Compression}
\label{sec:approach_compression}

We compress our model in two stages: first, iterative channel pruning is applied, followed by post-training quantization to 8-bit for both weights and activations.

\subsubsection{Channel Pruning.}
We propose structured channel pruning, which avoids the need for libraries with sparsity support on the target MCU and compresses the activation tensors as well to reduce RAM memory.
Each iteration, the 5\% filters with the lowest normalized $L_2$-norm, as proposed by Ophoff et al. \cite{ophoff2021investigating}, are removed, followed by a fine-tuning step.
When fine-tuning is finished, the weights with the lowest validation loss are selected to initialize the model for the next iteration.
To speed up the pruning process, we adopt the following early stopping criteria: given the validation loss of the current pruned model $L_{curr}$ and the validation loss of the unpruned model $L_{start}$, we skip or stop fine-tuning when $L_{curr} \le 1.03 \times L_{start}$ and immediately continue with the next iteration.

\subsubsection{Quantization.}
After channel pruning is completed, PTQ is applied to convert both the model's weights and activations to 8-bits. We use the Tensorflow Lite model converter for this task and provide a calibration set of 100 images to estimate the scales and zero-points of the activation tensors.

\section{Experiments}
\label{sec:experiments}
In the first section of our experiments (Section \ref{sec:results_comparison}), we compare our models accuracy-wise against the baseline,  which is the proprietary in-house software from Melexis. Section \ref{sec:results_bg_sub} studies the added value of our background subtraction method, Section \ref{sec:results_compress} elaborates on the compression ratios and accuracy losses and final Section \ref{sec:results_deploy} compares the time performance and resource occupation of our models, deployed with two different run-time frameworks on two different MCU targets.

\subsection{Comparison}
\label{sec:results_comparison}

In this section we compare the accuracy of our detection model against the accuracy of our baseline, the in-house developed detection software by Melexis, the manufacturer of the MLX90640.
Melexis's proprietary detection software is built and perfected by an in-house development team, over the course of two years. Their algorithm uses dynamic noise suppression, contrast and edge enhancement in pre-processing, and calculates a dynamic threshold to identify local maxima and minima after background subtraction. A multi-object tracker is used to further improve the accuracy.
Compared to a 90-degree or overhead camera view, detecting persons in video from a 45-degree camera angle turns out to be much more difficult. Therefore, accuracy results are reported separately for both views throughout the whole experiments section.

\begin{table}
    \caption{$F_1$-score accuracy results of our baseline -- Melexis's proprietary detection software -- compared to our models with and without compression applied.}
    \centering
    \begin{tabular}{|l|l|l|}
        \hline
        Model & $F_1$ 45-degree & $F_1$ 90-degree \\
        \hline
        Baseline & 64.31\% & 89.02\% \\
        Ours & {\bf 83.52\%} & {\bf 91.76\%} \\
        Ours compressed & 79.9\% & {\bf 91.62\%} \\
        \hline
    \end{tabular}
    \label{tab:result_comparison}
\end{table}

Table \ref{tab:result_comparison} presents the $F_1$-scores on the test set of our model in uncompressed and compressed state, and the baseline.
For both camera angles, our models outperform Melexis's software by 19.2\% and 2.7\% on the 45-degree and 90-degree test sets, respectively. In compressed state, where our model is heavily pruned and quantized to 8-bit, this still results in a performance boost of 15.6\% and 2.6\% on the 45-degree and 90-degree test sets, respectively, making our approach the preferred solution.

The Average-Precision (AP) metric, which is calculated from a Precision-Recall curve, is popular in most object detection papers, but requires a model to be able to produce a prediction probability for each produced bounding-box. Since Melexis's software is not capable of producing such probabilities, it does not make sense to use this metric here. We therefore prefer $F_1$-score instead, which is a percentile number produced from a single precision and a single recall value.
Selecting a single precision-recall point for a detector like ours requires selecting a single threshold value. We find the optimal threshold by maximizing the $F_1$-score through adjustment of the threshold value. We use a standard 0.5 overlap threshold for matching detections to ground-truth boxes and set the NMS overlap threshold to 0.3.

\subsection{Background Subtraction}
\label{sec:results_bg_sub}

\begin{figure}
    \centering
    \subfloat[]{\includegraphics[width=.5\textwidth]{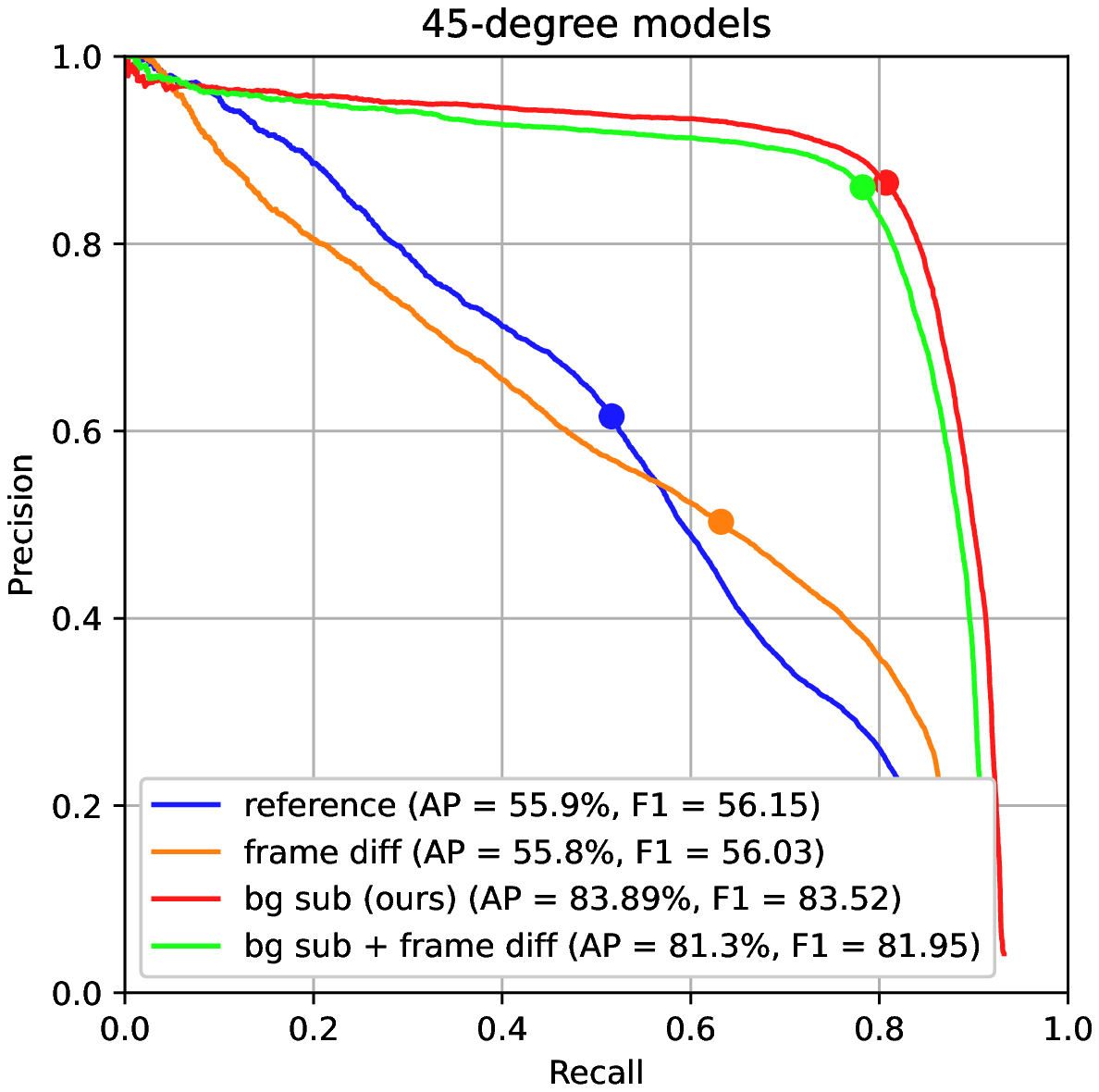}
    \label{fig:results_45_degree_models}}
    \subfloat[]{\includegraphics[width=.5\textwidth]{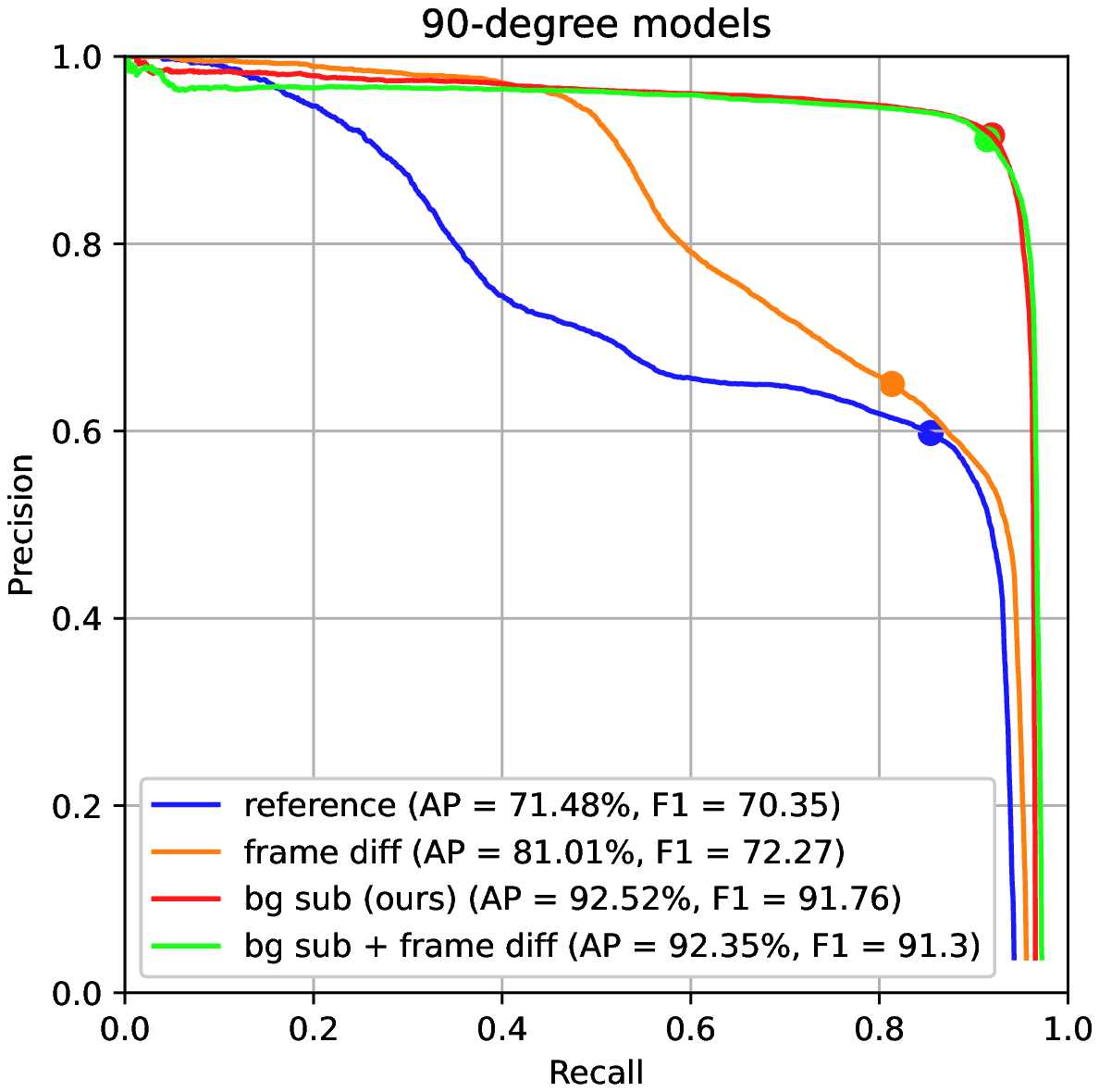}
    \label{fig:results_90_degree_models}}
    \caption{PR-curves of (a) 45-degree and (b) 90-degree models. For each curve, AP and $F_1$ scores are presented (best viewed in color).}
    \label{fig:results_models}
\end{figure}

This section presents results on the added value of our proposed background subtraction method. Figure \ref{fig:results_45_degree_models} and \ref{fig:results_90_degree_models} depict the PR-curve charts of a number of models tested on 45-degree and 90-degree labeled dataset images, respectively.
Models that are tested on the 90-degree test set are trained on the 90-degree training set, while model tested on the 45-degree test set are trained on both 90-degree and 45-degree training data.

\vspace{-2em}

Each chart presents results of four different models: (1) \textit{reference}, which is our model without background subtraction, (2) \textit{frame diff}, which is the reference model, but with an additional difference image on a second input channel, (3) \textit{bg sub}, which is our proposed model with background subtraction and (4) \textit{bg sub + frame diff}, which is our model with background subtraction and the additional difference image.
Inspired by the optical-flow input from Simonyan et al. \cite{simonyan2014two}, we experiment with a difference image, which is a computationally cheap alternative to optical-flow and a way to inject short-term motion information into the network. We construct the difference image by subtracting a previous image from the current image, where we set the frame stride between both images equal to five.
From our results, it can be concluded that this difference image seems to help improve the accuracy a little when applied to the \textit{reference} model on the 90-degree dataset. However, the added value of the difference image diminishes in comparison to our proposed background subtraction method, which boosts the accuracy by 28\% and 21\% AP on the 45-degree and 90-degree datasets, respectively. Since the background subtraction method eliminates the confusion between a person and an imposter background object, the network gets much less confusing samples during training, resulting in such a tremendous performance boost. The \textit{bg sub + frame diff} models however did not improve on the \textit{bg sub}, so the difference image is left out from our proposed method.

All models in Figure \ref{fig:results_models} are directly trained on our dataset from a random initialization for up to 50k iterations ($\le$3 hours on a GTX1080), using the SGD optimizer with a high weight decay $d=0.03$ to reduce overfitting.
The first 1000 iterations, a warm-up stage exponentially increases the learning rate up to the base learning rate $lr=0.001$. Subsequently, a reduce-on-plateau learning rate scheduler decays the learning rate with factor 10 whenever the validation loss did not drop in the previous 5000 iterations. For data augmentation, we use random contrast, brightness, horizontal and vertical flipping manipulations. 

\subsection{Model Compression}
\label{sec:results_compress}

The compression results of our 45-degree and 90-degree models are presented in Table \ref{tab:results_compression}. For each model, the AP and $F_1$ accuracies are given for the original model, the pruned model and the pruned + quantized model. We start with a 1.26M parameter model and prune that down to around a 10k parameters model, with negligible loss in accuracy for the 90-degree model and acceptable loss in accuracy for the 45-degree model. This results in pruning rates of up to $\div$136 and $\div$52 for the parameters and MACS, respectively. Since standard PTQ to 8-bit weights and activations hardly influences the accuracy, there is no need for QAT.

\begin{table}
    \caption{Accuracies, model size, number of Multiply-Accumulates and number of bits per weights/activation for our original, pruned and pruned + quantized models.}
    \centering
    \begin{tabular}{|l|l|l|l|l|l|}
        \hline
        Model & AP & $F_1$ & \#params & \#MAC & W/A \\
        \hline
  	    45-degree & 83.89 \% & 83.52 \%	& 1.26 M & 68.36 M & 32/32 \\
     	45-degree-pruned & 80.48 \% & 80.5 \% & 13.86 k ($\div$91) & 1.95 M ($\div$35) & 32/32 \\
        45-degree-pruned-quant & 79.47 \% &	79.9 \% & 13.86 k ($\div$91) & 1.95 M ($\div$35) & 8/8 \\
        \hline
        90-degree & 92.52 \% & 91.76 \% & 1.26 M & 68.36 M & 32/32 \\
  	    90-degree-pruned & 92.49 \% & 91.75 \% & 9.23 k ($\div$136) & 1.32 M ($\div$52) & 32/32 \\
  	    90-degree-pruned-quant & 92.29 \% & 91.62 \% & 9.23 k ($\div$136) & 1.32 M ($\div$52) & 8/8 \\
        \hline
    \end{tabular}
    \label{tab:results_compression}
\end{table}

Both models are pruned for up to 60 iterations on a GTX1080, which completes in about 1.5 days. We then select the model with the highest test accuracy that is close to 10k parameters.
The 45-degree and 90-degree models that we selected are pruned for 44 and 49 iterations, respectively.
Each fine-tuning step takes at most 10k iterations, which is 1/5\textsuperscript{th} of a regular training time. The learning rate is set to 0.0001 and lowered by a factor 10 after 5k iterations.

\subsection{Deployment}
\label{sec:results_deploy}

After pruning and quantization, we deploy both 45-degree and 90-degree models on two popular low-cost MCUs: (1) the STM32F746 with a Cortex-M7 core and (2) the STM32F407 with a Cortex-M4 core.
We compare the inference time and memory utilization between two open-source neural-network deployment frameworks for our application: Tensorflow Lite for microcontrollers (TFLite) \cite{tflite} and microTVM \cite{microtvm}. We configure both frameworks to use the CMSIS-NN \cite{lai2018cmsis} microkernel library as a backend, because it supports SIMD instructions that can execute four 8-bit multiply-accumulate operations in a single cycle, which significantly boosts the performance.
For our benchmark, we compile the generated C-code with all optimizations for speed enabled (\verb+GCC -O3+) and set the clock speeds of both platforms to their maximum, which is 216MHz for the STM32F746 and 168MHz for the STM32F407.

\begin{figure}
    \centering
    \includegraphics[width=.8\textwidth]{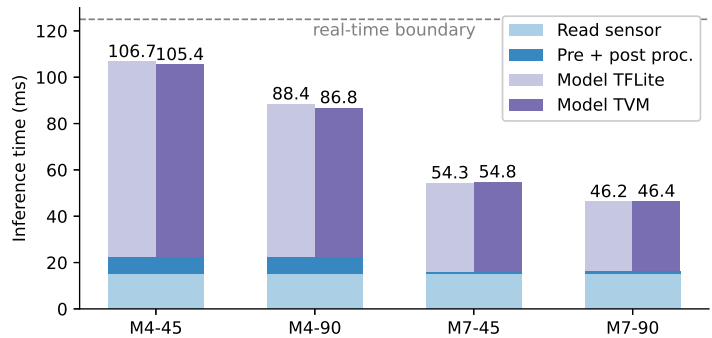}
    \caption{Inference times of both 45-degree and 90-degree models on an STM32F407 (M4) and an STM32F746 (M7) MCU, using either Tensorflow Lite (TFLite) or microTVM (TVM).}
    \label{fig:results_infer_times}
\end{figure}

Figure \ref{fig:results_infer_times} presents the inferences times of our models (\textit{Model TFLite/Model TVM}), sensor acquisition time (\textit{Read sensor}) and other processing (\textit{Pre + post proc}), which includes pixel temperature calculation, input normalization, yolo post-processing, NMS and background calculation. All our models are capable of running in real-time ($\ge$ 8 FPS) on both microcontrollers with sufficient headroom for doing other tasks.

\begin{figure}
    \centering
    \subfloat[RAM memory usage]{\includegraphics[width=.4\textwidth]{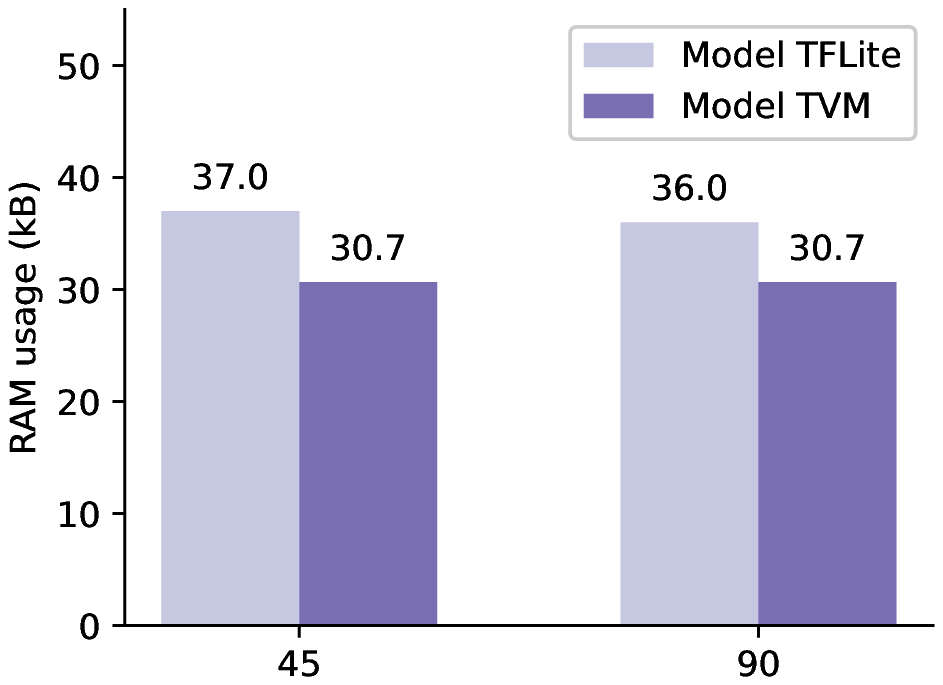}
    \label{fig:results_ram_usage}}
    \quad
    \subfloat[Flash memory usage]{\includegraphics[width=.4\textwidth]{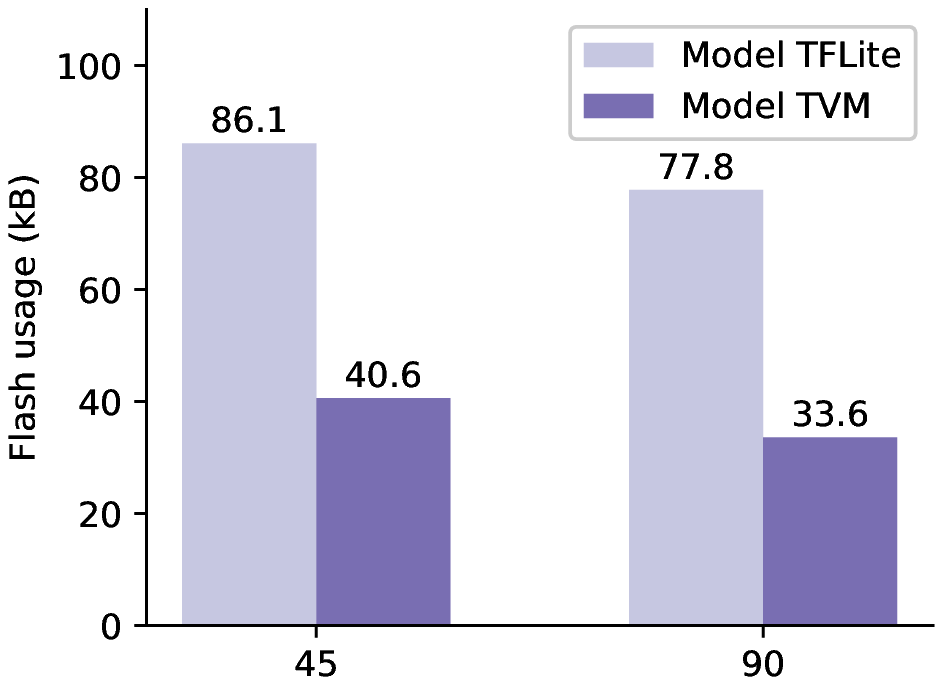}
    \label{fig:results_flash_usage}}
    \caption{RAM and flash memory usage statistics of both 45-degree and 90-degree models, measured when using either Tensorflow Lite (TFLite) or microTVM (TVM).}
\end{figure}

Figure \ref{fig:results_ram_usage} and \ref{fig:results_flash_usage} report the RAM memory and flash memory usage, respectively. The utilization of RAM memory in our microTVM experiments is noticeably lower compared to that of Tensorflow Lite, but the biggest difference can be seen in flash-memory usage, where microTVM uses less than half the size compared to Tensorflow Lite.
This is because Tensorflow Lite's model interpreter engine is a fixed size component that becomes significantly large when working with very small models like ours. In contrast, microTVM generates direct function calls to the CMSIS-NN micro kernels without the need for interpreter code, which tremendously reduces the flash usage. 
Note that the statistics only report the used resources of the model.

\section{Conclusion}
\label{sec:conclusion}

Although person detection is a wanted asset in many real-life applications, it is not always feasible due to (1) low-cost constraints and (2) privacy related issues if a regular camera is involved.

This work solves both problems by proposing a low-cost detection system in the order of magnitude of tens of dollars, based on a privacy-preserving ultra low-resolution thermal imager and a low-cost microcontroller. Even though it is extremely difficult to create an accurate deep-learning model that works with such limited resources, our compressed models achieve an accuracy of 79.9\% and 91.62\% (F1-score) on our 45-degree and 90-degree benchmark test, respectively, outperforming the standard person detection software of Melexis by a significant margin. We achieve our goal by proposing a smart background-subtraction mechanism that eliminates confusion between person-like objects and real persons, boosting our model's performance up to 28\%.
Our processing pipeline with 90-degree model is running at 46ms/image and 87ms/image on an STM32F746 and STM32F407, respectively, and only requires 34kB of flash memory and 31kB of RAM.

In future work, an additional spatio-temporal module could be added to potentially further improve the accuracy, or more extreme compression could be used to further down-size the model.

\subsubsection{Acknowledgements.}
This work is supported by VLAIO and Melexis via the AI@EDGE TETRA project.

%
%
%
\bibliographystyle{splncs04}
\bibliography{references}

\end{document}